\documentclass[lettersize,journal]{IEEEtran}
\usepackage{comment}
\usepackage{amsmath,amsfonts}
\usepackage{algorithmic}
\usepackage{algorithm}
\usepackage{mathtools}
\usepackage{array}
\usepackage{textcomp}
\usepackage{stfloats}
\usepackage{url}
\usepackage{verbatim}
\usepackage{comment}
\usepackage{graphicx}
\usepackage{subcaption} 
\usepackage{caption}

\usepackage{cite}
\usepackage{acronym}
\usepackage{booktabs}
\usepackage{mathtools}
\usepackage{comment}
\usepackage{multirow}
\usepackage{hhline}
\usepackage{adjustbox}
\usepackage{orcidlink}
\usepackage{svg}
\usepackage{hyperref}
\usepackage{comment}
\usepackage{placeins}
\usepackage{csquotes}
\hypersetup{
    colorlinks=true,
    linkcolor=blue,
    pdfborder={0 0 0},
    filecolor=blue,      
    urlcolor=black,
    citecolor=blue,
}

\begin{document}
\newcommand*\wideestimates{\mathrel{\widehat{=}}}
\newcommand*\estimates{\mathrel{\hat{=}}}

\title{SETTer: Sparse-Encoder Transformer for Long-term Multivariate Time Series Forecasting}

\author{
Abraham~Ezema\,$^{\orcidlink{0000-0002-9671-0925}}$\,,
Chijioke~Eze\,$^{\orcidlink{0000-0002-8545-0160}}$\,,
Ferdinanda~Ponci\ $^{\orcidlink{0000-0003-0431-9169}}$\,, and
Antonello~Monti\,$^{\orcidlink{0000-0003-1914-9801}}$\,

\thanks{Abraham Ezema is a PhD students with the Institute for Automation of Complex Power Systems, RWTH Aachen University, 52064 Aachen (e-mail: \{abraham.ezema@eonerc.rwth-aachen.de).}

\thanks{Chijike Eze is a PhD students with RWTH Aachen University, 52064 Aachen (e-mail: \{chijioke.eze@rwth-aachen.de).}

\thanks{Ferdinandi Ponci is a professor and Chair of the institute for Monitoring and Distributed Control for Power Systems Teaching and Research Area, RWTH Aachen University, 52064 Aachen (e-mail: fponci@eonerc.rwth-aachen.de).}

\thanks{Antonello Monti is a professor and Chair of the Institute for Automation of Complex Power Systems, RWTH Aachen University, 52064 Aachen and also Co-Director of Fraunhofer FIT, 53757 Sankt Augustin, Germany (e-mail: amonti@eonerc.rwth-aachen.de).}
}

\IEEEpubid{}

\maketitle

\begin{abstract}
Long-term multivariate time series plays a significant role in many application areas such as power systems, trading, etc. However, their accurate prediction is quite difficult for conventional forecasting methods as they often exhibit high dimensionality and complex relationships. Recent works show that transformer-based approaches are quite effective for long-term forecasting thanks to their attention mechanism. However, in the presence of complex high-dimensional inputs, they show evidence of oversmoothing, limited capacity, and opacity. To this end, this paper introduces SETTer, a transformer-based model that addresses these challenges by incorporating novel techniques for decoupled self-attention and hybrid masking. The proposed techniques enable SETTer to effectively capture the dominant short- and long-term patterns across the temporal and channel dimensions. In addition, we enrich the model layers with simple explainable structures that indicate the discriminative pattern of SETTer. We show that with a single-layer transformer architecture, SETTer can effectively model long-term dependencies in the presence of varying data complexities. Extensive experiments on real-word benchmark datasets for long-term multivariate time series forecasting demonstrate that SETTer outperforms state-of-the-art models in $88 \%$ of the scenarios.   

\end{abstract}


\section{Introduction}

\IEEEPARstart{M}{ultivariate} time series analysis plays a crucial role in decision support systems used in critical sectors such as energy, health, transportation, and commerce \cite{MVP_PAMI, forecast1, MVT_Transport2, MVP_Load2}, where they are the primary output of interest or input for other tasks. Generally, observations in multivariate time series data are ordered into temporal sequences, where each sequence represents multiple values of a variable (channel) over time. Long-term multivariate time series forecasting (LMTF), which predicts extended horizons given a historical input sequence, has recently been of notable interest due to its importance in decision-making processes of key service providers. For example, long-term energy generation forecasts can inform more precise structural planning for power system operators, enabling effective handling of wide-range seasonal variations in demand/supply profiles. This in turn can help the operators in capacity planning and long-term investment decisions, especially when several renewable energy resources (RES) are involved. 

\begin{figure}[t!]
    \centering
    \begin{subfigure}[b]{0.4\textwidth}
        \centering
        \includegraphics[width=3.1in]{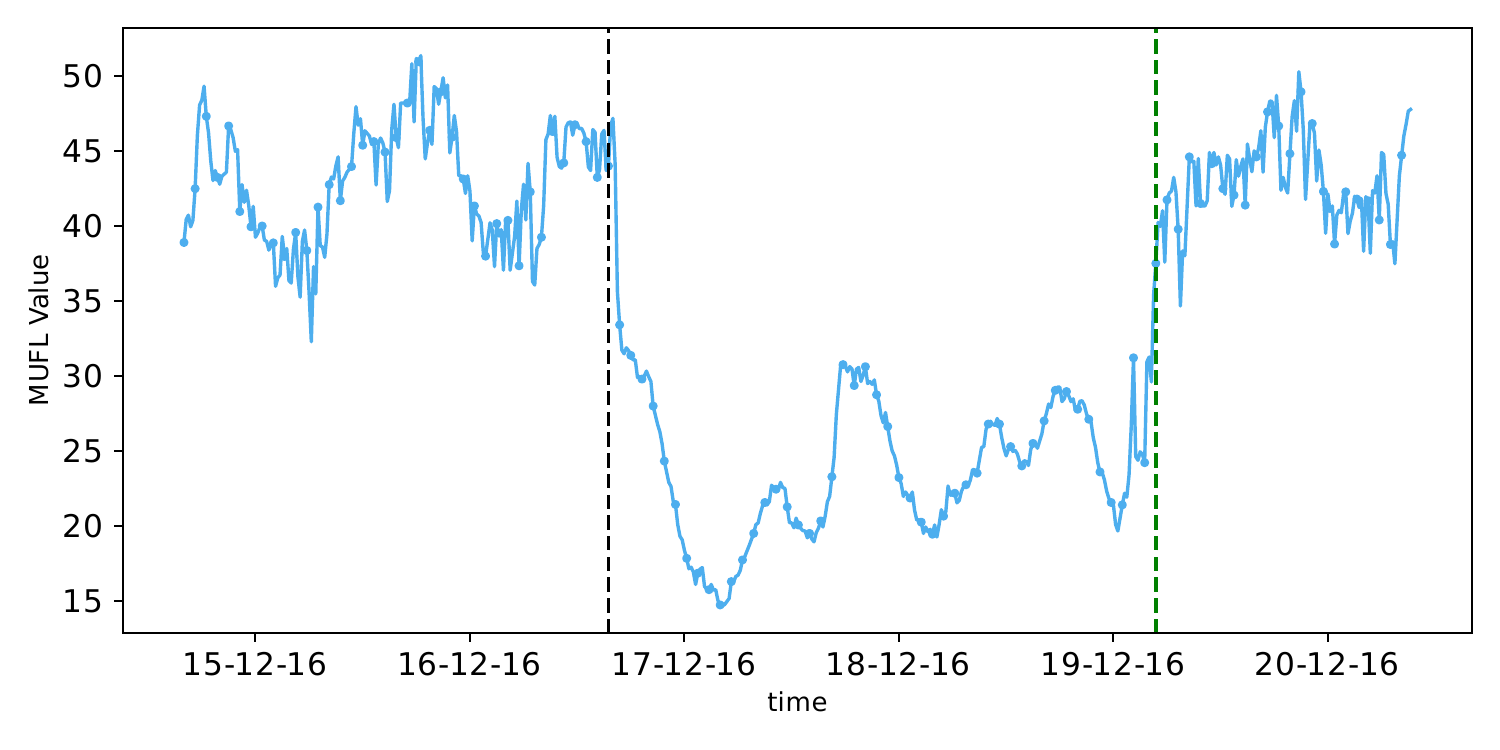}
        \captionsetup{justification=centering}
        \caption{Medium useful load (MUFL) of electric transformers}
        \label{fig: temporal patterns}
    \end{subfigure}
    \vspace{0.5em} 
    \hspace{0.7em}
    \begin{subfigure}[b]{0.17\textwidth}
        \includegraphics[width=\linewidth]{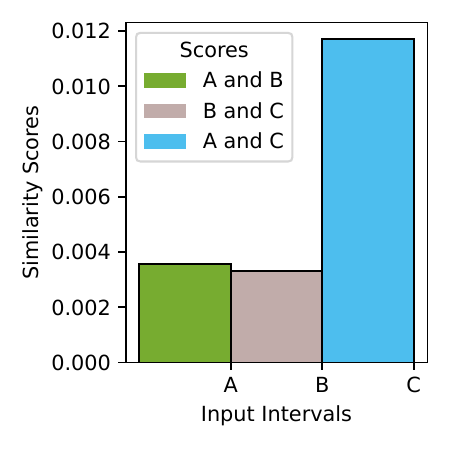}
        \captionsetup{justification=centering}
        \caption{Normalized euclidean similarity score.}
        \label{fig: interval similarity}
    \end{subfigure}
    \hspace{2.4em}
    \begin{subfigure}[b]{0.20\textwidth}
        \centering
        \includegraphics[width=1.4in]{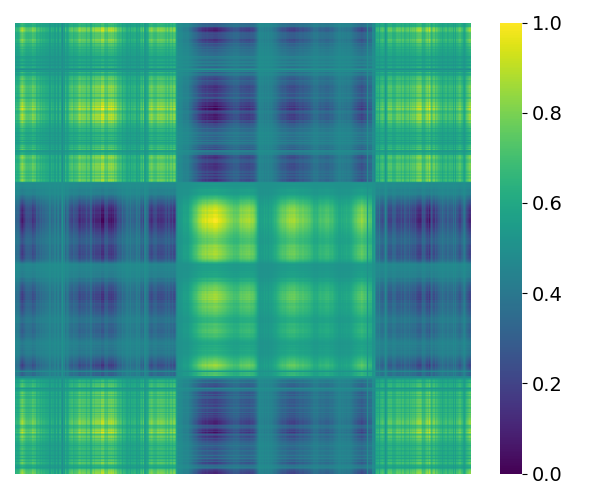}
        \vspace{0.28em}
        \captionsetup{justification=centering}
        \caption{Dependency Matrix of the intervals.}
        \label{fig: dependency matrix}
    \end{subfigure}

    \caption{Repetitive temporal pattern in data. In (a) and (b), the input sequence Interval B (middle) show low similarity with the target Interval C (right). In contrast, a larger input length Interval A (left and middle) show more similarity with the target. (c) illustrates high sparsity in the dependency matrix of the data, as well as the varying intra- and inter-interval dependency patterns.}
    \label{fig:tempProfile}
\end{figure}
Despite the benefits of LMTF, real-world applications can have strict accuracy requirements. For instance, an overestimation of RES generation can lead to excessive curtailment in power systems and result in the wastage of renewable energy. Moreover, real-world data tend to be complex, non-stationary, and exhibit dependencies across both temporal and channel dimensions. These dependencies demonstrate heterogeneity across domains, which makes it difficult for existing methods to effectively generalize \cite{TFB}. 

Most forecasting methods factor in the temporal dependence in data to capture the input relations over time \cite{ARMA, tempDep1, Dlinear}. Generally, data points in time series data exhibit short- and long-term temporal patterns that manifest as trends, seasonality, and cycles. Figure \ref{fig: temporal patterns} illustrates an example of this from energy data \cite{ETTdatasetRef} with recurring temporal patterns across different intervals, while Fig. \ref{fig: interval similarity} reveals the low similarity between close intervals and the high similarity between distant intervals. Therefore, a longer input length is necessary to capture these dynamics in the data. However, popular statistical methods such as Autoregressive Integrated Moving Average (ARIMA) and Gaussian processes \cite{ARMA, GPs} show  limitations in LMTF because the patterns are inconsistent with their underlying assumptions. Similarly, machine learning methods such as RNN has been applied to solve LMTF; however, they face the vanishing gradient problem which inhibits its ability to preserve long-term dependencies in the data \cite{LstmBadMem1}.  Recently, transformer-based methods have been widely adopted for LMTF due to their large receptive field via its attention mechanism which enables transformers to process long input sequences, regardless of length \cite{attentionallyouneed}. 

Nevertheless, a large input size does not necessarily result in a proportionally large effective receptive field. Particularly, although a transformer model processes information from all input positions, it may be predominantly short-term relations that have a significant influence on predictions \cite{effectiveReceptiveField}. This phenomenon leads to attention oversmoothing and attenuation of long-term relations in vanilla transformer models \cite{SAMFORMER, unstableTrans1, unstabletrans2}, which is attributable to rank collapse and short-term attention concentration. Existing improvements resort to modeling the dependencies implicitly by first enforcing a fixed local scope on inputs; i.e., patching, and then stacking layers to capture long-term dependencies between the predefined local scopes \cite{Patchtst,PDF, pathformer}. However, patch-based transformer formulations tend to have limited modeling capacity \cite{CD-CIComparisonForChannelDistriburionDrift}.

Similarly, block-sparse attention \cite{blocksparse1, blocksparseAttention, blocksparse3, blocksparse4} explores the addition of auxiliary structures to transformer encoders. Specifically, attention masks with strategically placed block regions that select only the most informative tokens in the attention matrix. This in turn shows the potential to preserve long-term dependencies. However, the application of fixed block regions can be too rigid for time series data, which show relation variations within short- and long-term intervals, as illustrated in Fig. \ref{fig: dependency matrix}. Consequently, developing more efficient sparse-encoder transformers is still an open research question, and remains mostly unexplored for time series forecasting.

Alternative methods propose integrating temporal dependence with cross-channel dependence to improve forecasting \cite{duet, crossformer, mcformer}. 
However, there are still significant challenges that affect channel dependence encoding for multivariate time series data. On the one hand, when there is strong inter-channel correlation, the methods that capture cross-channel dependence show remarkable performance, which highlights the relevance of cross-channel dependence for forecasting.  On the other hand, when there is weak inter-channel correlation, large channel size, and channel dependence shifts, the performance of the same methods degrade significantly \cite{TFB, CD-CIComparisonForChannelDistriburionDrift}. These suggest nuanced cross-channel dependence relevance in LMTF.

Motivated by the inherent challenges associated with multivariate time series encoding, this work explores the following research questions: First, how can transformers be augmented with auxiliary structures and mechanisms to effectively capture short- and long-term temporal dependencies in multivariate time series data without incurring huge computational cost? Second, can encoding methods effectively integrate cross-channel dependence in multivariate time series forecasting, especially considering its dual impact on forecasting performance? Lastly, how will such design choices contribute to improved model interpretability?
 
Based on the stated motivations, this work proposes a sparse-encoder transformer for long-term multivariate time series forecasting (SETTer), that systematically incorporates simple sparsification and adaptive masking mechanisms into vanilla transformer layers. In particular, SETTer introduces an alternative approach for temporal dependence modeling that uses a single sparse encoder layer to explicitly target short- and long-term dependencies. It levarages adaptive binarization to generate mask tokens from the input to control the encoder's span for long-term dependencies and captures short-term  dependencies with a fixed-span mask. In addition, SETTer uses an analogous masking technique to model cross-channel dependence and adds a channel-relevance attention block similar to the squeeze-and-excitation network \cite{squeezeandExcitation} to account for the dual impact of channel patterns. Specifically, the channel-relevance attention block modulates the global channel relationships in the input. Collectively, SETTer targets gains in both computational complexity and representational capacity of transformer-based architectures for LMTF  by strategically retaining informative structures in the attention mask.

In summary, our contributions include the following.
\begin{itemize}
    \item We propose a novel method for designing sparse encoder transformers for LMTF based on a hybrid masking scheme that is flexible to the inherent patterns in multivariate time series data. It targets the dominant short- and long-term temporal relations. 
    \item We propose a flexible channel dependency technique that adaptively weights the impact of channel-wise patterns in LMTF data and supplements the representational capacity and performance even for the largest channel dimensions.
    \item We demonstrate that this method is effective even with a simple transformer layer, essentially substituting deeper layers with simple auxiliary structures.
    \item We demonstrate that our design choices lead to more interpretable architectures that indicate the most favorable components in the temporal and channel dimensions that lead to forecasting improvements.
    \item We demonstrate the competitive performance of SETTer over state-of-the-art methods through extensive experiments conducted on 8 benchmark datasets for LTMF.
\end{itemize}

\section{Related Work}
This section reviews existing works that are most relevant to our approach. We focus on two key areas: (1) we outline different modeling approaches in transformer-based multivariate time series forecasting. These include channel-independent (CI) methods that model temporal dependence without inter-channel relationships, channel-dependent (CD) methods that integrate inter-channel relationships, channel clustering methods that combine CI and CD (2) transformer approaches that utilize block-sparse attention to process long sequences efficiently.

\subsection{Temporal and Channel Dependency Modeling}

Recent advances in multivariate time series forecasting have recognized the importance of modeling  temporal dependencies (relationships across time steps) and channel dependencies (relationships across different variables) to improve forecasting performance. For example, PatchTST \cite{Patchtst} introduced a foundational approach using subseries-level patches as input tokens and channel independence, to significantly improve long-term forecast precision while reducing computational complexity. Pathformer \cite{pathformer} extends patching to multi-scale modeling by introducing adaptive pathways that dynamically adjust the modeling process based on varying temporal dynamics, integrating both temporal resolution and temporal distance. Crossformer \cite{crossformer} introduced a Two-Stage Attention (TSA) mechanism that explicitly captures both cross-time and cross-dimension dependencies through a Dimension-Segment-Wise (DSW) embedding strategy, which transforms multivariate time series inputs into a 2D vector array, to preserve both temporal and channel-wise relationships. This specifically addresses the limitation of CI strategies that can lose important cross-channel information. 

To balance the benefits of CI with CD modeling more effectively, DUET \cite{duet} proposes a dual clustering framework that operates on both the temporal and channel dimensions. It employs a Temporal Clustering Module (TCM) to handle heterogeneous temporal patterns and a Channel Clustering Module (CCM) to capture complex inter-channel correlations.  However, fundamental issues remain with full-attention transformers in time series forecasting. For example, SAMformer \cite{SAMFORMER} identifies that transformers suffer from attention collapse and rank degeneration, requiring specialized optimization techniques to achieve competitive performance. This phenomenon is theoretically grounded in the finding that conventional self-attention networks converge doubly exponentially to rank-1 matrices \cite{unstableTrans1}, leading to token uniformity that undermines the expressive capacity of transformers. Despite these advances, existing CI methods tend to struggle when the channel dimension is large with significant interdependencies, while CD methods perform poorly when there is weak channel correlation.

\subsection{Block-Sparse Attention in Transformers}

The quadratic complexity of standard attention mechanisms has motivated the development of sparse attention patterns, particularly block-sparse attention approaches that strategically mask sections of the attention matrix to reduce computational costs while preserving long-range dependencies. Big Bird \cite{blocksparse1} introduced one of the basic block-sparse attention mechanisms, proposing a sparse attention pattern that combines local, global, and random connections. The approach maintains a few global tokens that attend to the entire sequence while using block-wise local attention and random connections to efficiently capture long-range dependencies, thus reducing the quadratic complexity to linear while maintaining the model's expressive capacity.

More recent developments have further refined these concepts. The Factorization Vision Transformer \cite{blocksparse3} addresses long-range dependency modeling by factoring attention computation into local windows, reducing computational costs while maintaining the ability to capture global relationships through hierarchical processing. To provide a more principled way to determine which blocks should be retained, XAttention \cite{blocksparseAttention} proposes an innovative approach that uses anti-diagonal scoring to identify important blocks in the attention matrix, leading to better accuracy-efficiency trade-offs compared to static sparse patterns. Additionally, the 2-D Transformer \cite{blocksparse4} extends sparse attention concepts to handle longer contexts with minimal memory requirements, demonstrating the scalability potential of well-designed sparse attention mechanisms to process extended sequences.

However, the resulting block patterns in these approaches are static for all input sequences, which makes them too rigid to capture more complex dependencies in time series data. Unlike natural language or image data where certain structural patterns are relatively consistent, time series data exhibit highly variable dependency patterns that change based on the underlying temporal dynamics and inter-variable relationships. In addition,  fixed block patterns cannot adapt to the varying correlation structures that characterize different time series, limiting their effectiveness in capturing the nuanced temporal and cross-channel dependencies essential for accurate multivariate forecasting.

Building upon these limitations, our proposed SETTer architecture introduces a channel-dependent adaptive masking mechanism that can dynamically adjust to the dependencies present in different time series while maintaining the computational benefits of sparse attention.

\section{Preliminaries}
\subsection{Problem Definition}
 We consider the multivariate forecasting framework: where given a collection of \emph{T} multivariate time series data  
$\boldsymbol{X}  \in \mathbb{R}^{T \times C } $, each time series, $X_{t} \in \boldsymbol{X}$ has a channel dimension \emph{C}. A slice of the data $\boldsymbol{X}_{t_i:t_j}$ is a sequence that represents the look-back of past observations of length $l=\left| t_j - t_i \right|$, and the horizon $\boldsymbol{X}_{t_{j + 1}: t_{j + h}}$, represents observations $h$-steps ahead from $\boldsymbol{X}_{t_j}$. The objective in forecasting is to find a mapping $f_{\boldsymbol{\Theta}}:\boldsymbol{X}_{t_i:t_j} \rightarrow \boldsymbol{Y}_{t_{j + 1}:t_{j + h} }$; i.e. from the look-back to the horizon, where $\boldsymbol{Y} \subseteq \boldsymbol{X}$, and $f$ is a sequence model that has a set of parameters $\boldsymbol{\Theta}$. 

\begin{figure*}[!t]
\centering
\subfloat[The attention and masking components of the sparse encoder.]{
\includegraphics[width=0.38\linewidth]{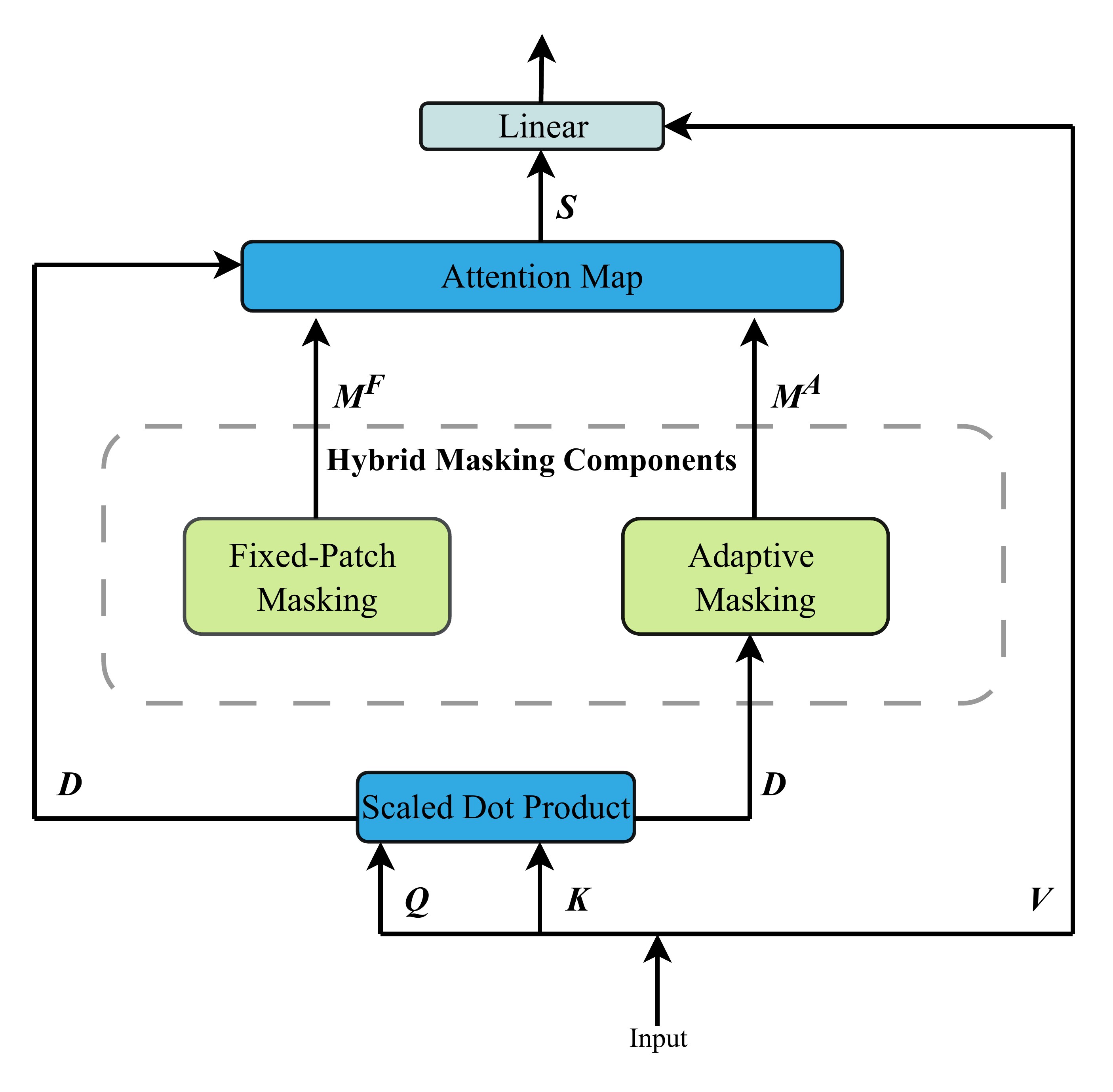}
\label{fig:AttentionComponents}
}
\hfill
\subfloat[The architecture of SETTer.]{
\includegraphics[width=0.52\linewidth]{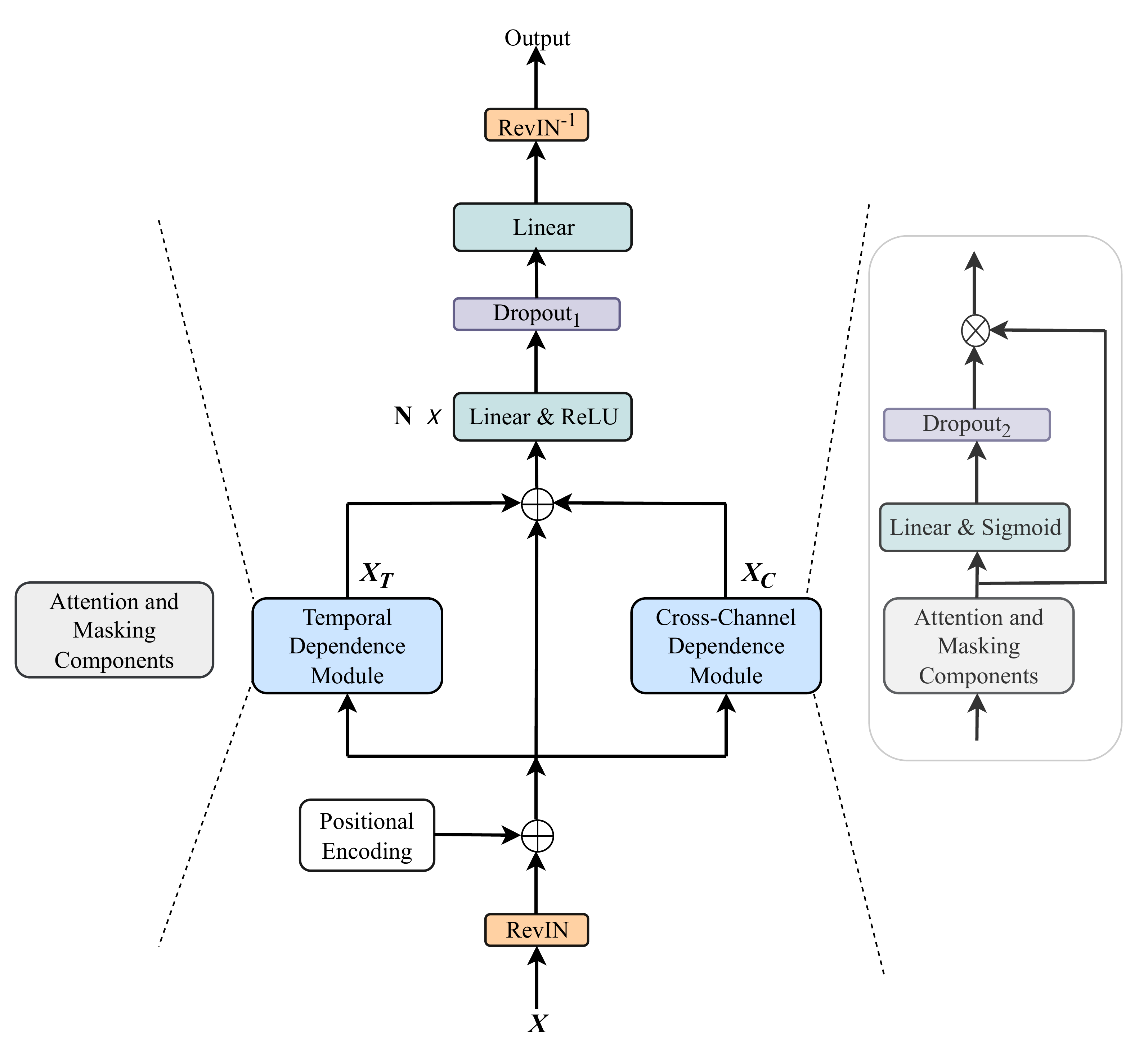}
\label{fig:fullarchitecture}
}
\caption{Overview of SETTer architecture. (a) The encoder computes the dependency matrix $\boldsymbol{D}$ from the keys ($\boldsymbol{K}$) and queries ($\boldsymbol{Q}$). Adaptive Masking generates $\boldsymbol{M}^A$ from  $\boldsymbol{D}$, while the Fixed-path Masking generates $\boldsymbol{M}^F$. The attention map combines the two masks ($\boldsymbol{M}^F$ and $\boldsymbol{M}^A$) to capture important relations in the attention weight matrix ($\boldsymbol{S}$). These form the building block of the two modules in SETTer architecture. (b) The Temporal and Cross-channel dependence modules use independent sparse encoders to capture temporal and channel relations in the input, respectively. The channel relations are further scaled before the residual connection, followed by a layer for non-linear temporal mixing and projection to the output dimension.}
\label{fig: SETTer architecture}
\end{figure*}

\subsection{Vanilla Transformer Architecture with Attention Mask}
\subsubsection*{Self-Attention}\label{section self-attention}
At the core of the transformer architecture \cite{attentionallyouneed} is the attention operation applied to the inputs. It maps an input to an output representation based on latent (tokens) similarities. The self-attention map computes similarities from  linear projections of input $\boldsymbol{X}$ into queries ($\boldsymbol{Q}$), keys ($\boldsymbol{K}$), and values ($\boldsymbol{V}$) using weights $\boldsymbol{W}_q \in \mathbb{R}^{C \times d_m}$, $\boldsymbol{W}_k \in \mathbb{R}^{C \times d_m}$, and $\boldsymbol{W}_v \in \mathbb{R}^{C \times d_m}$. The resulting projections are defined as $\boldsymbol{Q}=\boldsymbol{X}\boldsymbol{W}_q $, $\boldsymbol{K}=\boldsymbol{X}\boldsymbol{W}_k $, and $\boldsymbol{V}=\boldsymbol{X}\boldsymbol{W}_v $, where $d_m$ is the embedding dimension. A scaled dot product of $\boldsymbol{Q}$ and $\boldsymbol{K}$ results in a cross-covariance matrix\footnote{It is the cross-covariance matrix if $\boldsymbol{X}$ is centered} that gives a similarity score between all rows $X_{t_i}, X_{t_j} \in \boldsymbol{X} $. Next, a  $\mathrm{Softmax}$ function normalizes the cross-covariance matrix, and makes the outcome right stochastic, such that each row represents a probability distribution that highlights larger similarities and de-emphasizes smaller ones. The self-attention operations are defined as follows:
\begin{align}
\begin{split}\label{equation att-cross-cov}
    \boldsymbol{D} = \frac{\boldsymbol{Q}\boldsymbol{K}^\top}{\tau}; \quad \boldsymbol{D} \in \mathbb{R}^{T \times T }
\end{split}\\
\begin{split}\label{equation att-map}
   \boldsymbol{S} = \mathrm{Softmax} \left(\boldsymbol{D} \odot \boldsymbol{M} \right); \quad \boldsymbol{S} \in \mathbb{R}^{T \times T }
\end{split}\\
\begin{split}\label{equation att-out}
    \boldsymbol{O} =  \boldsymbol{S}\boldsymbol{V}\boldsymbol{W}_o; \quad \boldsymbol{O} \in \mathbb{R}^{T \times C}
\end{split}
\end{align}
where, $\boldsymbol{D}$ is the cross-covariance (dependency) matrix,  $\tau$ is the $\mathrm{Softmax}$ temperature scaling on $\boldsymbol{D}$. Higher $\tau$ values smoothings the output distribution in \ref{equation att-map}; in contrast, smaller values result in sharper differences in the output. In this work, we employ a learnable temperature scaling from \cite{ViTforsmallDataset}, to favor sharper distribution values. 
$\boldsymbol{M} \in \mathbb{F}_2^{T \times T}$ denotes the attention mask and is composed of the values $0$ and $1$, which represent the mask and the visible token, respectively. $\boldsymbol{M}$ enforces a sparse self-attention encoder. Specifically, the Hadamard product in \ref{equation att-map} associates each element in $\boldsymbol{D}$ with an attention mask token. Consequently,  the mask token serves to block elements of $\boldsymbol{D}$ from the $\mathrm{Softmax}$ computation. Conversely, the visible token selects elements for the $\mathrm{Softmax}$ normalization step. Sparse self-attention exploits these conditioning on $\boldsymbol{D}$ to constrain the relevance of row relations in the attention map. $\boldsymbol{S}$ is the resultant masked attention weight matrix, which scales the values ($\boldsymbol{V}$) to generate the transformer output in \ref{equation att-out}; and $\boldsymbol{W}_o \in \mathbb{R}^{d_m \times C}$ represents the projection weight of the output.

\subsubsection{Relationship between cross-covariance matrix and time series analysis}
$\boldsymbol{D}$ encodes the dependencies across observations in all time steps, such that for all $ D_{ii} \text{, } D_{ij} \in \boldsymbol{D}$ representing the diagonal and off-diagonal relations: $D_{ii} \estimates  \mathrm{var}(X_i, X_i)$, while $D_{ij}  \estimates \mathrm{cov}(X_i, X_j)$\footnote{The \enquote{estimates} operator is used for the variance and covariance due the different projection weights for queries and keys.}, and $X_i ,X_j \in \boldsymbol{X}$ represent the input values at different time steps. Thus,  $\boldsymbol{D}$ encapsulates long-term dependencies under the conditions that $j \gg i$ and $j \ll i$, and short-term dependencies when $j \not\gg i$  and $j \not\ll i$. Therefore, we exploit these relationships in $\boldsymbol{D}$ to initially isolate key components using $\boldsymbol{M}$, and subsequently emphasize them with a custom attention map. The key components consist of temporal dependencies (short-term and long-term), channel variance, and cross-channel dependencies.

\section{METHODOLOGY}
\subsection{SETTer Structure Overview}
Figure \ref{fig:fullarchitecture} illustrates the overall architecture of SETTer. The first block (RevIN) \cite{revin} normalizes the input ($\boldsymbol{X}$) to mitigate the impact of distribution shifts between training and test data. Next, a learnable positional embedding (PE) similar to that in \cite{BERT} adds information about the sequential order of the input. 
This is followed by two independent transformer layers with single-head backbones, specifically adapted to encode temporal dependencies via the temporal dependence module (TDM) and cross-channel dependencies with the cross-channel dependence module (CCDM).  The proposed modules are based on equations \ref{equation att-cross-cov}-\ref{equation att-out} with two important modifications.  First, each transformer backbone is augmented with auxiliary structures that specifically target short- and long-term dependencies for each token, while effectively isolating the influence of other relations. These in turn preserve the magnitude of important dependencies in the input. Second, we add structures that further scale the channel dependency to improve the representational capacity, especially in the presence of a large channel dimensionality. Subsequently, a residual connection adds the normalized input to the outcomes of TDM and CCDM. The following multilayer perceptron (MLP) block with rectified linear unit ($\mathrm{ReLU}$) activation \cite{ReLU} and a subsequent $\mathrm{Dropout}$ layer \cite{dropout} combine the residual result to capture non-linear dependencies. Finally, a linear predictor forecasts the output values, and the final output ($\boldsymbol{\hat{Y}}$) is the de-normalized value of the predictions.
The subsequent sections provide comprehensive details of the different components of SETTer.

\subsection{Temporal Dependence Module (TDM)} \label{section TDM}
\begin{figure*}[!t]
\centering
\subfloat[Fixed-patch mask generation for the dense partition.]{
\includegraphics[width=0.13\linewidth]{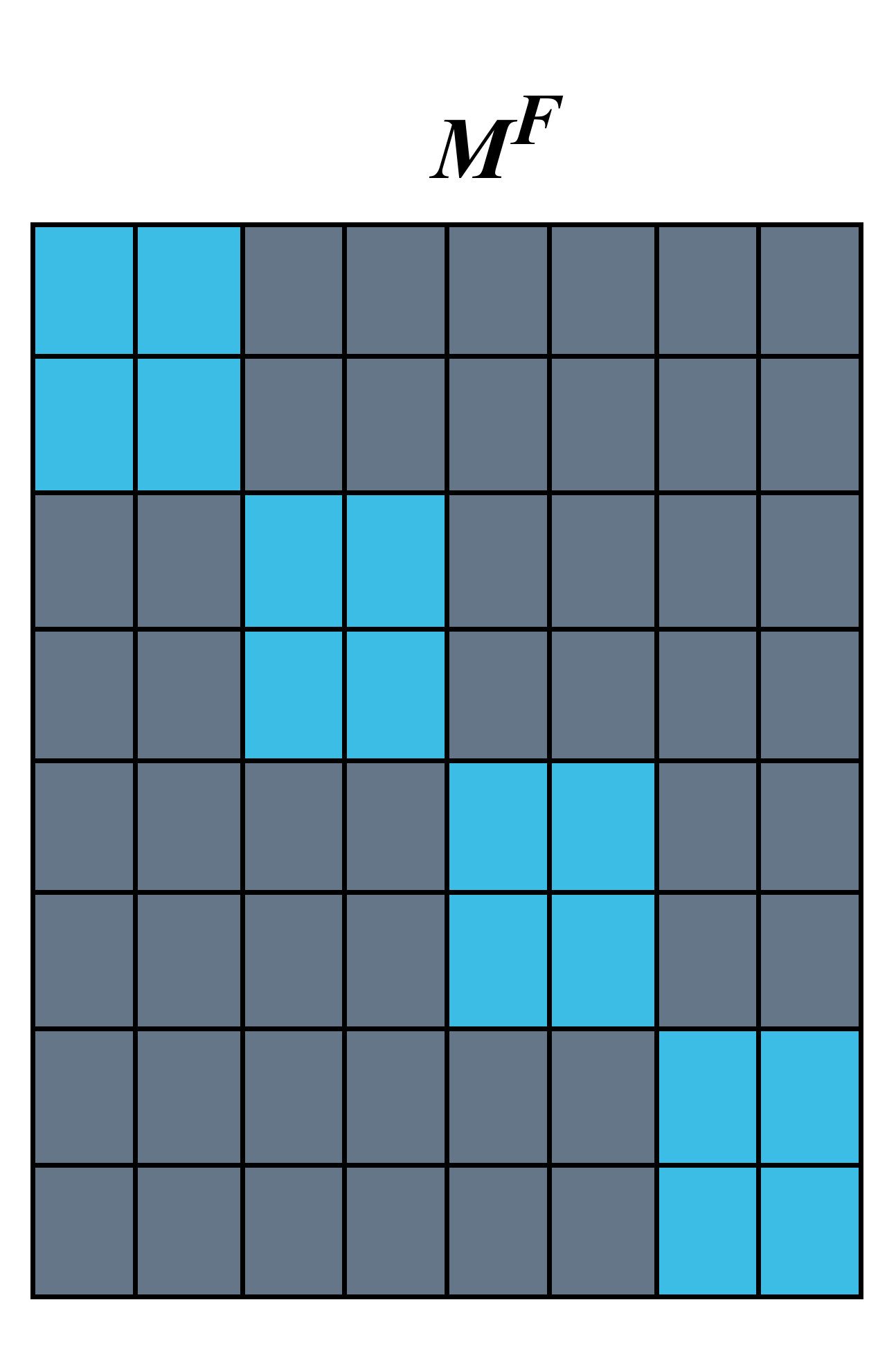}
\label{fig:fixed}
}
\hfill
\subfloat[Adaptive mask generation for the sparse partition.]{
\includegraphics[width=0.405\linewidth]{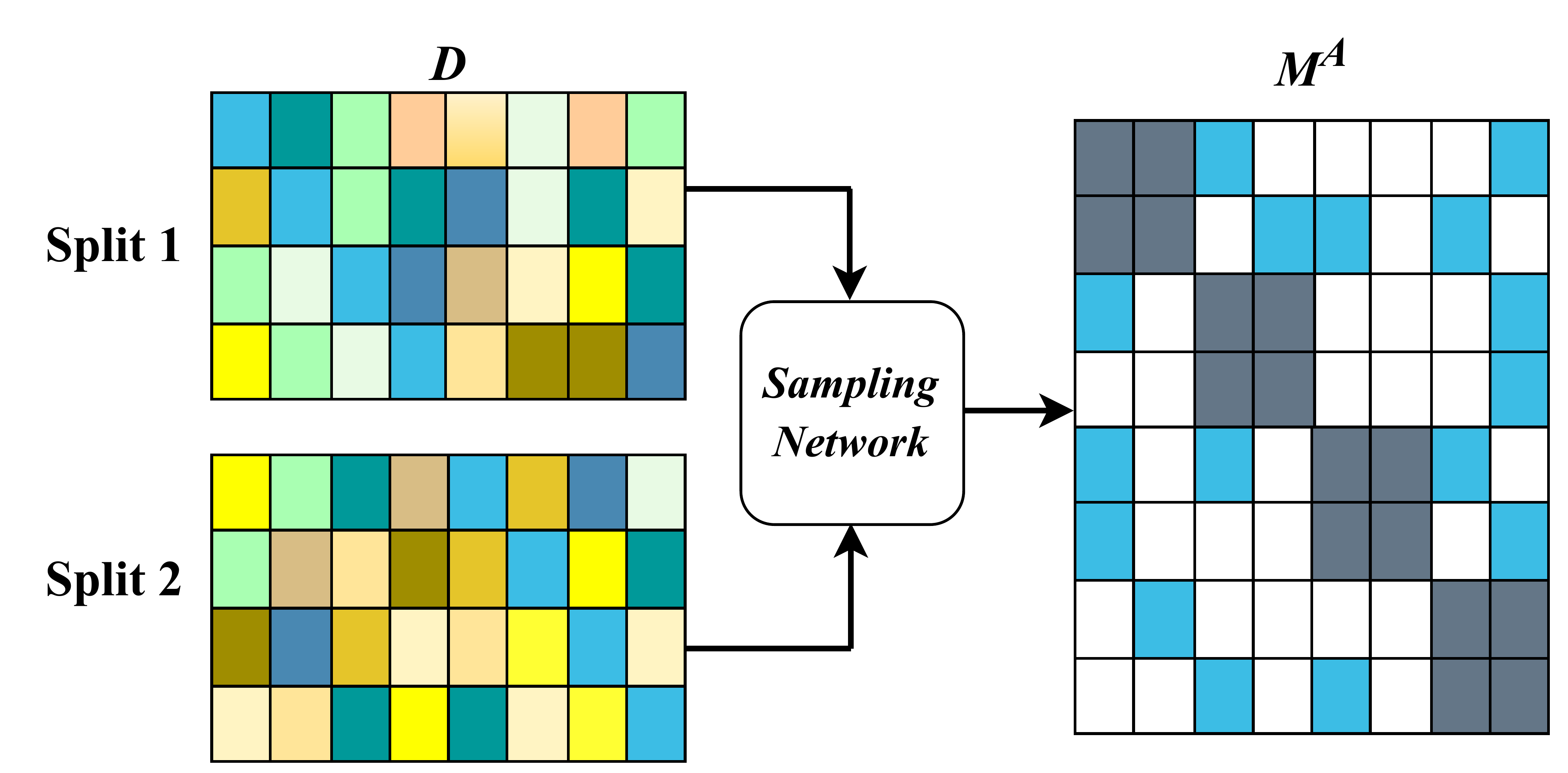}
\label{fig:adaptive partition1}
}
\hfill
\subfloat[Combination of the dense and sparse partition in the attention mask.]{
\includegraphics[width=0.3\linewidth]{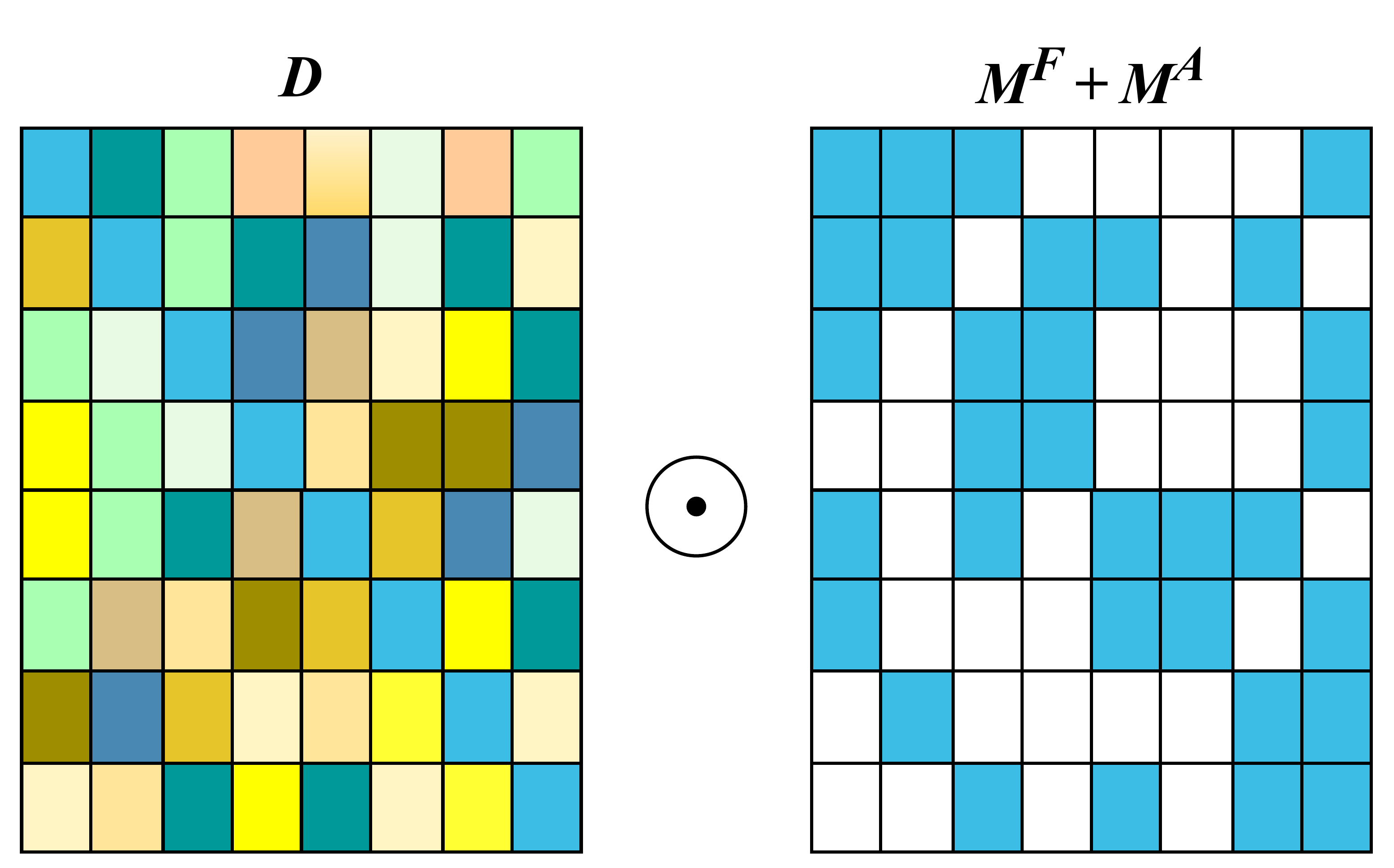}
\vspace{0.3em}
\label{fig:hybrid partition}
}
\caption{The hybrid masking mechanism in SETTer. (a) The dense partition is in blue (representing visible token), with patch width  $k = 2$, while the sparse partition is dark (representing mask token). The diagonal patches in  $\boldsymbol{M}^F$ also comprise the diagonal elements and limit the attention computations to short-term relations for each token. (b) The sampling network updates $\boldsymbol{M}^A$  using \emph{split} samples of $\boldsymbol{D}$. The sampling network highlights relevant long-term relations in $\boldsymbol{M}^A$  (blue) and the white positions are the other relations in the sparse partition it blocks. (c) The mask partitions are combined to select regions in $\boldsymbol{D}$.}
\label{fig: mask generation}
\end{figure*}

The standard attention in \ref{equation att-map} only considers the pointwise similarity between observations at each time step. In particular, due to softmax normalization, the influence of significant relations can be smoothed  out for large-dimensional inputs. Further, this approach does not account for the potential varying influence that nearby time steps have, in contrast to farther time steps.
Therefore, to account for such variations, we design a sparse transformer encoder that is enriched with a hybrid masking mechanism. It employs a fixed and adaptive masking scheme to simultaneously facilitate the propagation of short- and long-term relations in the attention mechanism, as illustrated in Fig. \ref{fig:AttentionComponents}. The proposed encoder attention map is defined as follows:
\begin{align}
   \begin{split}\label{equation: temp self-att}
   \boldsymbol{S} =  \sum_{\boldsymbol{M} \in \{\boldsymbol{M}^F, \boldsymbol{M}^A\} }\mathrm{Softmax} \left(\boldsymbol{D} \odot\boldsymbol{M} \right)
   \end{split}\\
   \begin{split}\label{equation: TCM output}
    \boldsymbol{X}_T =  \boldsymbol{S}\boldsymbol{V}\boldsymbol{W}_o; \quad \boldsymbol{X}_t \in \mathbb{R}^{T \times C}
    \end{split}
\end{align}
where, $\boldsymbol{S}$ is the attention weight matrix. Note that due to the summation in \ref{equation: temp self-att}, $\boldsymbol{S}$ is no longer right-stochastic. $\boldsymbol{D}$, $\boldsymbol{V}$ and $\boldsymbol{W}_o$ represent the dependency matrix, values, and output weights as described in Section \ref{section self-attention}; while $\boldsymbol{M}^F$ and $\boldsymbol{M}^A$ denote fixed and adaptive masks, respectively.
In principle, we constrain the dependency matrix ($\boldsymbol{D}$) of the encoder into dense and sparse partitions and adopt a 2-step attention map to decouple the impact of the relations in the partitions.
The dense partition retains short-term dependencies by applying $\boldsymbol{M}^F$ to block the influence of long-term relations in the first attention step. Conversely, in the second step, the sparse partition preserves long-term dependencies using $\boldsymbol{M}^A$ to select other temporal relations outside $\boldsymbol{M}^F$ that are relevant for forecasting.

\subsubsection{Fixed-Patch Masking (FPM)} We define two operations on an arbitrary mask $\boldsymbol{M}^F \in \mathbb{F}_2^{T \times T}$. The first operation is  a projection of rectangular patches $\lceil{\frac{T}{k}}\rceil \cdot \lceil{\frac{T}{k}}\rceil$ of size $k$ on $\boldsymbol{M}^F$; and then the reassignment of diagonal patches to visible tokens and off-diagonal patches to mask tokens as illustrated in  Fig. \ref{fig:fixed}. $\boldsymbol{M}^F$ defines non-overlapping patches, such that the diagonal patches also cover the diagonal tokens. 
Formally, for an arbitrary square matrix $\boldsymbol{M}^F = \sum_{i, j}^{T} \boldsymbol{M}^F_{ij}E_iE_j^\top$, where $E_i$ and $E_j$ are basis vectors, $\boldsymbol{M}^F_{ij}$ denotes the element of the matrix at position ($i$, $j$).
Hence, a projection $P$ on $\boldsymbol{M}^F$ gives the intersection of all elements $\boldsymbol{M}^F_{ij}$  with patches, such that  \[ \begin{split} P_{mn} = \{ \boldsymbol{M}^F_{ij} \mid \forall_{i,j \in T} \quad {}& m \cdot k \leq i < \min \bigl((m + 1) \cdot k, T\bigr)  \wedge \\ & n \cdot k \leq j < \min\bigl((n + 1) \cdot k, T\bigr),\\
& m < \lceil{\frac{T}{k}}\rceil \wedge  n < \lceil{\frac{T}{k}}\rceil \} .\end{split} \]

Given the projections, function $\varphi(.)$ reassigns the patch tokens as follows: 
\begin{equation}
    \begin{split}
    \varphi(P_{mn}) = \left \{
      \begin{aligned}
        &1, && \text{if}\ m = n \\
        &0, && \text{otherwise} 
      \end{aligned} \right.
    \end{split}
    \end{equation}
where, the off-diagonal patches constitute the sparse partition.

\subsubsection{Adaptive Masking (AM)}
We adopt an adaptive mask sampling network ($\mathcal{A_{SN}}$) to modify the sparse partition mask $\boldsymbol{M}^A$, whereby for each input token, it adaptively generates a corresponding visible or mask token. The premise is to capture global patterns in the input sequence by encoding dependencies over an optimal subset of input tokens. Moreover, the token generation approach should be informative and optimal, so that it improves the loss function and is differentiable in the $\mathcal{A_{SN}}$ parameters. To this end, we use the dependency matrix $\boldsymbol{D}$ as input to $\mathcal{A_{SN}}$, ensuring that the AM outcome is derived exclusively from the relative similarity between the input tokens. Note that parameterization on $\boldsymbol{D}$ will constrain the transformer to generate visible mask tokens for only the most relevant relations. This eliminates the need for fixed data-dependent thresholds. The $\mathcal{A_{SN}}(.)$ operations are divided into four steps as follows:

\begin{align}
\begin{split}\label{split and FFN operations}
    \boldsymbol{D}^{S_0} = \mathrm{Split(\boldsymbol{D})},\quad \boldsymbol{D}^{S_1} = \mathrm{NN}_{\theta}(\boldsymbol{D}^{S_0})
\end{split}\\
\begin{split}\label{concat operation}
    \boldsymbol{D}^{S_2} = \mathrm{Concat}(\boldsymbol{D}^{S_1})
\end{split}\\
\begin{split}\label{sigmoid operation}
    \boldsymbol{M}^{A} = \sigma(\beta \cdot \boldsymbol{D}^{S_2})
\end{split}\\
\begin{split}\label{partition operation}
    \boldsymbol{M}^{A} = (1 - \boldsymbol{M}^{F}) \odot \boldsymbol{M}^{A}  .
\end{split}
\end{align}

Equations \ref{split and FFN operations}--\ref{sigmoid operation} constitute the sparsification steps based on token importance estimation. In general, it generates masks from slices of $\boldsymbol{D}$, rather than from complete-pass computations, which can lead to overfitting (as observed from our experiments). Figure \ref{fig:adaptive partition1} illustrates an example of the adaptive masking operations. Note that, for a split number $n_t$, $\mathrm{Split(.)}$ divides $\boldsymbol{D}$  into $\lceil{\frac{d}{n_t}}\rceil$ slices, where $d = \mathrm{rank}(\boldsymbol{D})$ is the row rank of $\boldsymbol{D}$. Each slice has length $s_l = \lfloor{\frac{d}{n_t}}\rfloor$, with $i = 0$ and $j = s_l$ indicating the initial and terminal indices of these slices, respectively. Thus,
\[ \boldsymbol{D}^{S_0} = [ \boldsymbol{D}_{i:j}, \boldsymbol{D}_{i + s_l:j + s_l}, \ldots, \boldsymbol{D}_{i + (n_t -1 )\cdot s_l:j + (n_t -1 )\cdot s_l}  ] \]
 is a collection of non-overlapping slices that share a common fully-connected sparsification network ($\mathrm{NN}_{\theta}$). $\mathrm{NN}_{\theta}$ adaptively tunes the threshold for the boolean function ($\sigma$) in \ref{sigmoid operation} through its parameters $\theta \in \mathbb{R}^{d \times d}$. We utilize the $\mathrm{Sigmoid}$ function as $\sigma$ to add nonlinearity, whereby temperature $\beta$ controls its smoothness to produce visible and mask tokens, respectively. That is, as $\beta \rightarrow \infty$, $\sigma_{\beta} : \boldsymbol{M}^A \in \mathbb{R}^{d \times d} \rightarrow \boldsymbol{M}^A \in  \mathbb{F}_2 ^{d \times d}$, inducing a hard saturation in $\sigma$. Equation \ref{partition operation} masks the dense partition already covered by $\boldsymbol{M}^F$, as illustrated by the dark patches in Fig. \ref{fig:adaptive partition1}. Figure \ref{fig:fixed}--\ref{fig:hybrid partition} is an example of the resultant hybrid masking partitions defined in \ref{equation: temp self-att}. Together, these features of $\mathcal{A_{SN}}$ make TDM differentiable and suitable for end-to-end training.

 \subsection{Cross-Channel Dependence Module(CCDM)}
\begin{figure}[ht]
     \begin{flushleft}
     \includegraphics[width=3.53in]{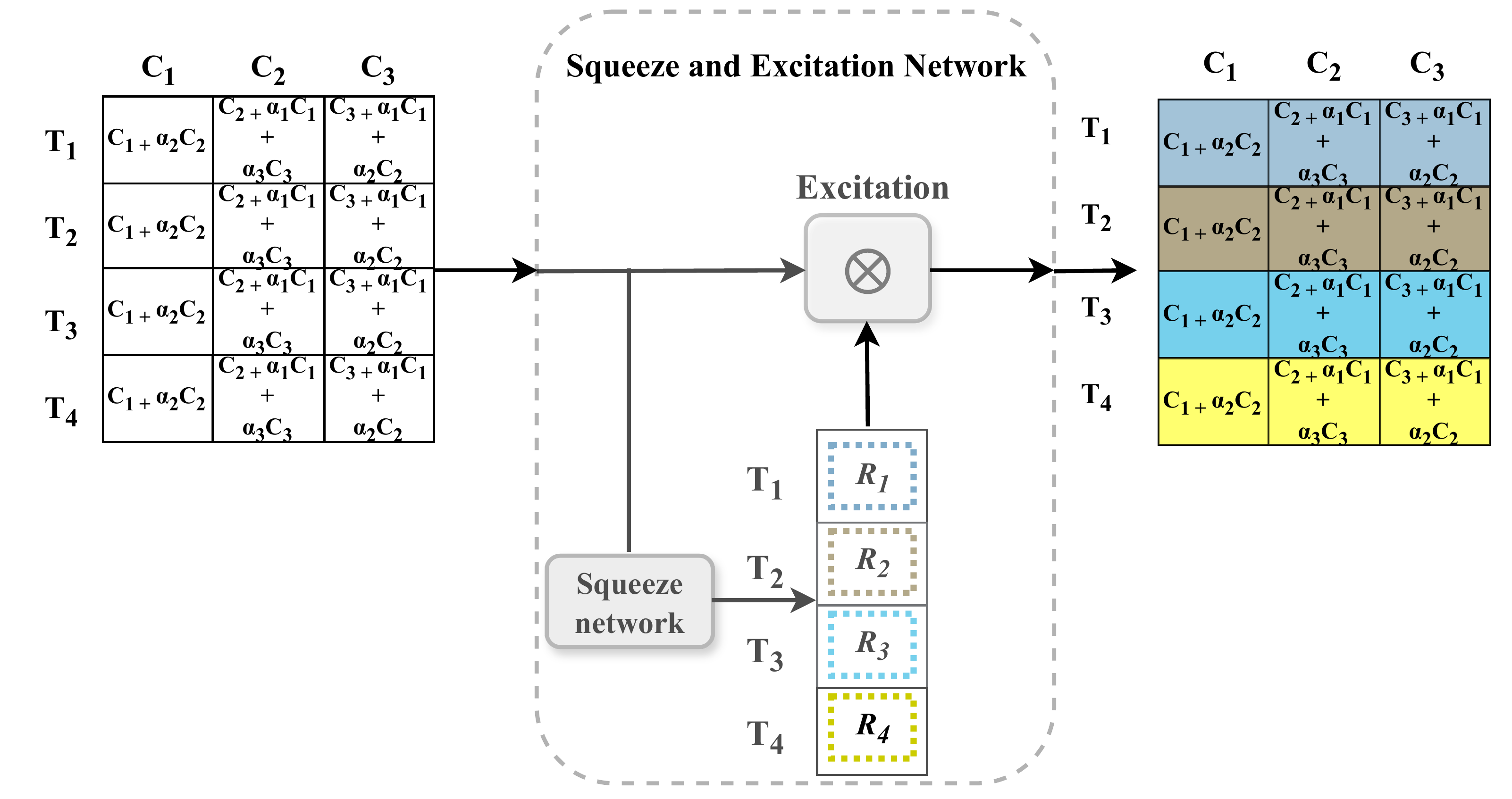}
     \end{flushleft}
    \caption{The overall structure of the cross-channel dependency module (follows a squeeze-and-excitation network). The channel-wise dependency have common channel relations in all time steps, these relations are scaled in each time step by the attention scores from the squeeze network to modulate their effect on the output.}
    \label{fig: channel_summarize}
\end{figure}

 The present strategies for cross-channel dependence result in global channel interdependencies and representations. As a result, they do not account for channel dependency shifts in multivariate time series data. Thus, we adopt a channel dependency strategy that consists of channel-wise self-attention and channel-relevance attention. The self-attention component captures cross-channel dependency using transformer architecture with masking to highlight self- and inter-channel relations. Additionally, channel-relevance attention modulates the influence of channel-wise relations for each time step. Further, due to the residual connection in Fig. \ref{fig:fullarchitecture} that adds CCDM and TDM outcomes, channel-relevance attention dynamically attenuates the cross-channel dependence when it is noisy, but propagates it when it supplements the temporal patterns for forecasting. The channel self-attention operations are defined as follows: 
 \begin{align}
\begin{split}\label{equation: cross-channel dependence}
    \boldsymbol{D}^{'} \hspace{1.8ex} = \frac{\boldsymbol{X}^\top\boldsymbol{W}^{'}_{q}\boldsymbol{W}_{k}^{'\top}\boldsymbol{X}}{\tau{'}}; \quad \boldsymbol{D^{'}} \in \mathbb{R}^{C \times C }
\end{split}\\
\begin{split}\label{equation: generate adaptive mask for channel}
     \boldsymbol{M}^{'A} = \mathcal{A_{SN}}(\boldsymbol{D}^{'}, \hspace{0.5ex} \boldsymbol{M}^{'F}); \quad  \boldsymbol{M}^{'A} \in \mathbb{R}^{C \times C }
\end{split}\\
\begin{split}\label{equation: channel attention weight}
   \boldsymbol{S}^{'} \hspace{2.4ex}  =  \sum_{\boldsymbol{M}^{'} \in \{\boldsymbol{M}^{'F}, \boldsymbol{M}^{'A}\} }\mathrm{Softmax} (\boldsymbol{D}^{'}  \odot \boldsymbol{M}^{'} )
\end{split}\\
\begin{split} \label{equation: channel attention output}
\boldsymbol{O^{'}} \hspace{1.8ex}  =  \boldsymbol{S}^{'}\boldsymbol{X}^\top\boldsymbol{W}^{'}_{v}\boldsymbol{W}^{'}_{o}; \quad \boldsymbol{O}^{'} \in \mathbb{R}^{C \times T}
\end{split}
\end{align}
where, $\boldsymbol{D}^{'}$ and $\boldsymbol{S}^{'}$ are the channel dependency and attention weight matrices, respectively.  Note that due to the summation in \ref{equation: channel attention weight}, $\boldsymbol{S}^{'}$ is not right stochastic. $\boldsymbol{W}^{'}_{q} \in \mathbb{R}^{T \times d^{'}_m}$, $\boldsymbol{W}^{'}_{k} \in \mathbb{R}^{T \times d^{'}_m}$,  $\boldsymbol{W}^{'}_{v} \in \mathbb{R}^{T \times d^{'}_m}$, and  $\boldsymbol{W}^{'}_{o} \in \mathbb{R}^{d^{'}_m \times T}$ represent the linear transformation weights of the attention map, and $d^{'}_m$ is the dimension of the channel-wise transformer model. $\boldsymbol{M}^{'F} = I_C$, analogous to $\boldsymbol{M}^{F}$ highlights the self-relation of channel tokens and isolates the sparse partition in $\mathcal{A_{SN}}$ for adaptive channel masking. Similarly, $\boldsymbol{M}^{'A}$ is a select subset of informative channel relations. Equations \ref{equation: channel attention weight}--\ref{equation: channel attention output} notably retain the following structures in $ \boldsymbol{S}^{'}$ and $\boldsymbol{O^{'}}$: for each channel $c_i \in C$, the two-step attention map first concentrates the highest attention weight on $c_i$, so that $\boldsymbol{S}^{'}_{ii} = 1$ and in the next step distributes the attention weights between other channels $c_j \neq c_i$. Therefore, the encoder combines a strong channel-wise self-relevance with weighted inter-channel influence, as illustrated in  Fig. \ref{fig: channel_summarize}. 

In contrast, channel-relevance attention applies soft attention to modulate cross-channel dependence, which in turn improves the representational capacity of CCDM, as illustrated in Fig. \ref{fig: channel_summarize}. The channel-relevance attention is represented as follows:
\begin{align}
\begin{split}\label{equation: channel relevance attention 1}
   R_c = \sigma(\mathrm{NN}_{\theta_2}(\boldsymbol{O}^{'\top}));\quad R_c \in \mathbb{R}^T
\end{split}\\
\begin{split}\label{equation: channel relevance attention 2}
    R_c = \mathrm{Dropout}_2(R_c) 
\end{split}\\
\begin{split}\label{equation: channel relevance attention 3}
    \boldsymbol{X}_C = R_c \odot \boldsymbol{O}^{'\top}; \quad \boldsymbol{X}_C \in \mathbb{R}^{T \times C}
\end{split}
\end{align}
where, $\sigma$ is the $\mathrm{Sigmoid}$ function, $\mathrm{NN}_{\theta_2}$ is a layer of FCN with parameter $\theta_2 \in \mathbb{R}^{C \times 1}$, and $\mathrm{Dropout}_2(.) $ denotes a dropout layer \cite{dropout}. Equation \ref{equation: channel relevance attention 1} is a \emph{squeeze} operation on the cross-channel dependence. It first maps each position to a distinct representation and then assigns a weight to it. Equation \ref{equation: channel relevance attention 3} is an \emph{excitation} operation that scales the channel dependency at each position.

\subsection{Regularization}
\subsubsection{Random Mask Drop} In addition to Dropout \cite{dropout} and RevIN \cite{revin}, we also randomly drop temporal mask patches in $\boldsymbol{M}^F$, starting with the last dense diagonal patch. A total of $n_d = \lfloor{\lceil{\frac{l}{k}}\rceil \cdot p_d}\rfloor$ patches constitute the drop candidates, where $p_d$ is the drop percentage. The drop scheme reassigns the elements of the $n_d$ patches to mask tokens and adds them to the sparse partition for adaptive masking. This in turn causes  $n_d$ sparse partitions to be interspersed with the remaining $ \lceil{\frac{l}{k}}\rceil - n_d$ dense partitions on the diagonal. This approach introduces more flexibility to the temporal masking structure of SETTer.

\subsubsection{Sharpness-Aware Minimization (SAM)}
We employ SAM \cite{originalSAM} to alleviate the training instability in transformers \cite{SAMFORMER, transformerdifficulttraining}. SAM formulates the training phase as a min-max optimization problem to obtain model parameters ($\Theta$) that have neighbors with low loss $ \mathcal{L} (\cdot)$. SAM training is defined as
\[  \min_{\Theta}\mathcal{L}_{SAM}(\Theta) = \min_{\Theta} \max_{ \left \|\epsilon\right \| \leq \rho_s} \mathcal{L}(\Theta + \epsilon), \]
where $\rho_s$ is the size of the neighborhood.

\begin{table*}[!ht]
    \centering
    \caption{Multivariate forecasting results with forecasting horizons $h \in \{96,192,336,720\}$ and look-back $l = 512$. \textbf{Best} results are in bold, \underline{second best}  are underlined.} 
    \label{tab:multivariate-forecasting-results2}
    \tiny
    \renewcommand{\arraystretch}{1.15} 
    \begin{adjustbox}{width=\textwidth,center}
    \begin{tabular}{ll|cc|cc|cc|cc|cc|cc|cc|cc|cc|cc|cc}
    \toprule
    \multicolumn{2}{c|}{\textbf{Models}} & \multicolumn{2}{c|}{\textbf{SETTer}} & \multicolumn{2}{c|}{\textbf{DUET}} & \multicolumn{2}{c|}{\textbf{PDF}} & \multicolumn{2}{c|}{\textbf{SAMformer}} & \multicolumn{2}{c|}{\textbf{iTransformer}} & \multicolumn{2}{c|}{\textbf{Pathformer}} & \multicolumn{2}{c|}{\textbf{FITS}} & \multicolumn{2}{c|}{\textbf{PatchTST}} & \multicolumn{2}{c|}{\textbf{Crossformer}} & \multicolumn{2}{c}{\textbf{DLinear}} \\
    \multicolumn{2}{c|}{} & \multicolumn{2}{c|}{(ours)} & \multicolumn{2}{c|}{(2025)} & \multicolumn{2}{c|}{(2024)} & \multicolumn{2}{c|}{(2024)} & \multicolumn{2}{c|}{(2024)} & \multicolumn{2}{c|}{(2024)} & \multicolumn{2}{c|}{(2024)} & \multicolumn{2}{c|}{(2023)} & \multicolumn{2}{c|}{(2023)} & \multicolumn{2}{c}{(2023)} \\
    \cmidrule(lr){3-4} \cmidrule(lr){5-6} \cmidrule(lr){7-8} \cmidrule(lr){9-10} \cmidrule(lr){11-12} \cmidrule(lr){13-14} \cmidrule(lr){15-16} \cmidrule(lr){17-18} \cmidrule(lr){19-20} \cmidrule(lr){21-22}
    \multicolumn{2}{c|}{Metrics} & MSE & MAE  & MSE & MAE  & MSE & MAE  & MSE & MAE  & MSE & MAE  & MSE & MAE  & MSE & MAE  & MSE & MAE  & MSE & MAE  & MSE & MAE \\
    \midrule
    \multirow{4}{*}{ETTh1} & 96 & \textbf{0.350} & 0.386 & \underline{0.352} & \textbf{0.384} & 0.360 & 0.391 & 0.362 & \underline{0.385} & 0.386 & 0.405 & 0.372 & 0.392 & 0.376 & 0.396 & 0.377 & 0.397 & 0.411 & 0.435 & 0.379 & 0.403 \\
    & 192 & \textbf{0.389} & \underline{0.410} & 0.398 & \textbf{0.409} & \underline{0.392} &  0.414 & 0.405 & \underline{0.410} & 0.424 & 0.440 & 0.408 & 0.415 & 0.400 & 0.418 & 0.409 & 0.425 & 0.409 & 0.438 & 0.408 & 0.419 \\
    & 336 & \textbf{0.409} & \textbf{0.424} & \underline{0.414} & 0.426 & 0.418 & 0.435 & 0.427 & \underline{0.425} & 0.449 & 0.460 & 0.438 & 0.434 & 0.419 & 0.435 & 0.431 & 0.444 & 0.433 & 0.457 & 0.440 & 0.440 \\
    & 720 & \textbf{0.420} & \textbf{0.447} & \underline{0.429} & 0.455 & 0.456 & 0.462 & 0.430 & \underline{0.448} & 0.495 & 0.487 & 0.450 & 0.463 & 0.435 &  0.458 & 0.457 & 0.477 & 0.501 & 0.514 & 0.471 & 0.493 \\
    \hhline{--|--|--|--|--|--|--|--|--|--|--}
    \multirow{4}{*}{ETTh2} & 96 & \textbf{0.264} & \textbf{0.330} & \underline{0.270} & \underline{0.336} & 0.276 & 0.341 & 0.290 & 0.340 & 0.297 & 0.348 & 0.279 & \underline{0.336} & 0.277 & 0.345 & 0.274 & 0.337 & 0.728 & 0.603 & 0.300 & 0.364 \\
    & 192 & \textbf{0.322} & \textbf{0.369} & 0.332 & \underline{0.374} & 0.339 & 0.382 & 0.385 & 0.401 & 0.372 & 0.403 & 0.345 & 0.380 & \underline{0.331} & 0.379 & 0.348 & 0.384 & 0.723 & 0.607 & 0.387 & 0.423 \\
    & 336 & \textbf{0.346} & \textbf{0.392} & 0.353 & 0.397 & 0.374 & 0.406 & 0.423 & 0.436 & 0.388 & 0.417 & 0.378 & 0.408 & \underline{0.350} & \underline{0.396} & 0.377 & 0.416 & 0.740 & 0.628 & 0.490 & 0.487 \\
    & 720 & 0.386 & 0.426 & \textbf{0.382} & \textbf{0.425} & 0.398 & 0.433 & 0.405 & 0.440 & 0.424 & 0.444 & 0.437 & 0.455 & \textbf{0.382} & \textbf{0.425} & 0.406 & 0.441 & 1.386 & 0.882 & 0.704 & 0.597 \\
    \hhline{--|--|--|--|--|--|--|--|--|--|--}
    \multirow{4}{*}{ETTm1} & 96 & \textbf{0.275} & \textbf{0.328} & \underline{0.279} & \underline{0.333} & 0.286 & 0.340 & 0.291 & 0.336 & 0.300 & 0.353 & 0.290 & 0.335 & 0.303 & 0.345 & 0.289 & 0.343 & 0.314 & 0.367 & 0.300 & 0.345 \\
    & 192 & \textbf{0.317} & \textbf{0.353} & \underline{0.320} & \underline{0.358} & 0.321 & 0.364 & 0.328 & 0.360 & 0.341 & 0.380 & 0.337 & 0.363 & 0.337 & 0.365 & 0.329 & 0.368 & 0.374 & 0.410 & 0.336 & 0.366 \\
    & 336 & \textbf{0.347} & \textbf{0.373} & \underline{0.348} & \underline{0.377} & 0.354 & 0.383 & 0.364 & 0.381 & 0.374 & 0.396 & 0.374 & 0.384 & 0.368 & 0.384 & 0.362 & 0.390 & 0.413 & 0.432 & 0.367 & 0.386 \\
    & 720 &  \textbf{0.404} &  \textbf{0.404} & \underline{0.405} & \underline{0.408} & 0.408 & 0.415 & 0.426 & 0.414 & 0.429 & 0.430 & 0.428 & 0.416 & 0.420 & 0.413 & 0.416 & 0.423 & 0.753 & 0.613 & 0.419 & 0.416 \\
    \hhline{--|--|--|--|--|--|--|--|--|--|--}
    \multirow{4}{*}{ETTm2} & 96 & \textbf{0.158} & \textbf{0.245} & \underline{0.161} & \underline{0.248} & 0.163 & 0.251 & 0.165 & 0.282 & 0.175 & 0.266 & 0.164 & 0.250 & 0.165 & 0.254 & 0.165 & 0.255 & 0.296 & 0.391 & 0.164 & 0.255 \\
    & 192 & \textbf{0.211} & \textbf{0.282} & \underline{0.214} & \underline{0.286} & 0.219 & 0.290 & 0.233 & 0.304 & 0.242 & 0.312 & 0.219 & 0.288 & 0.219 & 0.291 & 0.221 & 0.293 & 0.369 & 0.416 & 0.224 & 0.304 \\
    & 336 & \textbf{0.261} & \textbf{0.316} & \underline{0.267} & 0.321 & 0.269 & 0.330 & 0.286 & 0.335 & 0.282 & 0.337 & 0.267 & \underline{0.319} & 0.272 & 0.326 & 0.276 & 0.327 & 0.588 & 0.600 & 0.277 & 0.337 \\
    & 720 & \textbf{0.345} & \textbf{0.372} & \underline{0.348} & \underline{0.374} & 0.349 & 0.382 & 0.366 & 0.379 & 0.375 & 0.394 & 0.361 & 0.377 & 0.359 & 0.381 & 0.362 & 0.381 & 0.750 & 0.612 & 0.371 & 0.401 \\
    \hhline{--|--|--|--|--|--|--|--|--|--|--}
    \multirow{4}{*}{Weather} & 96 & \textbf{0.141} & \textbf{0.184} & 0.146 & \underline{0.191} & 0.147 & 0.196 & 0.174 & 0.215 & 0.157 & 0.207 & 0.148 & 0.195 & 0.172 & 0.225 & 0.149 & 0.196 & \underline{0.143} & 0.210 & 0.170 & 0.230 \\
    & 192 & \textbf{0.185} & \textbf{0.226} & \underline{0.188} & \underline{0.231} & 0.193 & 0.240 & 0.216 & 0.258 & 0.200 & 0.248 & 0.191 & 0.235 & 0.215 & 0.261 & 0.191 & 0.239 & 0.198 & 0.260 & 0.216 & 0.273 \\
    & 336 & \underline{0.236} & \textbf{0.266} & \textbf{0.234} & \underline{0.268} & 0.245 & 0.280 & 0.263 & 0.290 & 0.252 & 0.287 & 0.243 & 0.274 & 0.261 & 0.295 & 0.242 & 0.279 & 0.258 & 0.314 & 0.258 & 0.307 \\
    & 720 & \underline{0.307} & \textbf{0.319} & \textbf{0.305} & \textbf{0.319} & 0.323 & 0.334 & 0.329 & 0.338 & 0.320 & 0.336 & 0.318 & 0.326 & 0.326 & 0.341 & 0.312 & 0.330 & 0.335 & 0.385 & 0.323 & 0.362 \\
    \hhline{--|--|--|--|--|--|--|--|--|--|--}
    \multirow{4}{*}{Electricity} & 96 & \textbf{0.128} & \underline{0.221} & \textbf{0.128} & \textbf{0.219} & \textbf{0.128} & 0.222 & 0.139 & 0.234 & 0.134 & 0.230 & 0.135 & 0.222 & 0.139 & 0.237 & 0.143 & 0.247 & 0.134 & 0.231 & 0.140 & 0.237 \\
    & 192 & \textbf{0.144} & \textbf{0.235} & \underline{0.145} & \textbf{0.235} & 0.147 & 0.242 & 0.153 & 0.246 & 0.154 & 0.250 & 0.157 & 0.253 & 0.154 & 0.250 & 0.158 & 0.260 & 0.146 & 0.243 & 0.154 & 0.251 \\
    & 336 & \textbf{0.160} & \textbf{0.254} & \underline{0.163} & \underline{0.255} & 0.165 & 0.260 & 0.168 & 0.260 & 0.169 & 0.265 & 0.170 & 0.267 & 0.170 & 0.268 & 0.168 & 0.267 & 0.165 & 0.264 & 0.169 & 0.268 \\
    & 720 & \textbf{0.189} & \textbf{0.277} & \underline{0.193} & \underline{0.281} & 0.199 & 0.289 & 0.208 & 0.294 & 0.194 & 0.288 & 0.211 & 0.302 & 0.212 & 0.304 & 0.214 & 0.307 & 0.237 & 0.314 & 0.204 & 0.301 \\
    \hhline{--|--|--|--|--|--|--|--|--|--|--}
    \multirow{4}{*}{Solar} & 96 & \textbf{0.167} & 0.210 & \underline{0.169} & \textbf{0.195} & 0.181 & 0.247 & 0.207 & 0.230 & 0.190 & 0.244 & 0.218 & 0.235 & 0.208 & 0.255 & 0.170 & 0.234 & 0.183 & \underline{0.208} & 0.199 & 0.265 \\
    & 192 & \textbf{0.184} & \underline{0.220} & \underline{0.187} & \textbf{0.207} & 0.200 & 0.259 & 0.235 & 0.247 & 0.193 & 0.257 & 0.196 & \underline{0.220} & 0.229 & 0.267 & 0.204 & 0.302 & 0.208 & 0.226 & 0.220 & 0.282 \\
    & 336 & \textbf{0.190} & 0.226 & 0.199 & \textbf{0.213} & 0.208 & 0.269 & 0.249 & 0.251 & 0.203 & 0.266 & \underline{0.195} & \underline{0.220} & 0.241 & 0.273 & 0.212 & 0.293 & 0.212 & 0.239 & 0.234 & 0.295 \\
    & 720 & \underline{0.204} & \underline{0.237} & \textbf{0.202} & \textbf{0.216} & 0.212 & 0.275 & 0.256 & 0.256 & 0.223 & 0.281 & 0.208 & \underline{0.237} & 0.248 & 0.277 & 0.215 & 0.261 & 0.215 & 0.256 & 0.243 & 0.301 \\
    \hhline{--|--|--|--|--|--|--|--|--|--|--}
    \multirow{4}{*}{Traffic} & 96 & \textbf{0.359} & \textbf{0.237} & \underline{0.360} & \underline{0.238} & 0.368 & 0.252 & 0.420 & 0.283 & 0.363 & 0.265 & 0.384 & 0.250 & 0.400 & 0.280 & 0.370 & 0.262 & 0.526 & 0.288 & 0.395 & 0.275 \\
    & 192 & \textbf{0.378} & \textbf{0.247} & \underline{0.383} & \underline{0.249} & \underline{0.382} & 0.261 & 0.426 & 0.279 & 0.384 & 0.273 & 0.405 & 0.257 & 0.412 & 0.288 & 0.386 & 0.269 & 0.503 & 0.263 & 0.407 & 0.280 \\
    & 336 &  \underline{0.394} & \textbf{0.257} & 0.395 & \underline{0.259} & \textbf{0.393} & 0.268 & 0.433 & 0.281 & 0.396 & 0.277 & 0.424 & 0.265 & 0.426 & 0.301 & 0.396 & 0.275 & 0.505 & 0.276 & 0.417 & 0.286 \\
    & 720 & \textbf{0.435} & \underline{0.282} & \textbf{0.435} & \textbf{0.278} & 0.438 & 0.297 & 0.469 & 0.300 & 0.445 & 0.308 & 0.452 & 0.283 & 0.478 & 0.339 & \textbf{0.435} & 0.295 & 0.552 & 0.301 & 0.454 & 0.308 \\
    \midrule
    \multicolumn{2}{l|}{1st Count} & 27 & 23 & 6 & 11 & 2 & 0 & 0 & 0 & 0 & 0 & 0 & 0 & 1 & 1 & 1 & 0 & 0 & 0 & 0 & 0 \\
    \bottomrule
    \end{tabular}
    \end{adjustbox}
    \end{table*}

\section{Evaluation}
\subsection{Setup}
\subsubsection{Dataset}
We evaluate the performance of SETTer on eight popular datasets for LMTF benchmarking in the literature \cite{Patchtst, TFB}. The datasets comprise real-world observations from energy (ETTh1, ETTh2, ETTm1, ETTm2, Solar, Electricity), transportation (Traffic) and meteorology (Weather), and are publicly available on \cite{ETTdatasetRef ,autoformer, TFB}. Table \ref{tab:benchmarkDatasets} summarizes the properties of the datasets, additional details can be found in \cite{autoformer}. The Preprocessing operations on the datasets, such as splitting and normalization, follow the settings in \cite{SAMFORMER, duet}.

\begin{table}[H]
\tiny
\centering
\renewcommand{\arraystretch}{1.2}
\caption{Statistics of benchmark datasets.}
\label{tab:benchmarkDatasets}
\begin{tabular}{c | c | c | c c c c}
\hline
\textbf{Datasets} & \textbf{ETTh1/h2} & \textbf{ETTm1/m2} & \textbf{Weather} & \textbf{Electricity} & \textbf{Solar}  & \textbf{Traffic}\\ \hline
Channels & 7 & 7 & 21 & 321 & 137 & 862 \\
Timesteps & 17420 & 69680 & 52696 & 26304 & 52560 & 17544 \\
Granularity & 1 hour & 15 minutes & 10 minutes & 1 hour & 10 minutes & 1 hour\\
\hline
\end{tabular}
\end{table}

\subsubsection{Baselines}
We choose several state-of-the-art models for LMTF as our baselines. These include CI patch-based transformers (PatchTST \cite{Patchtst}, PathFormer \cite{pathformer}, PDF \cite{PDF}), CD transformers (iTransformer \cite{itransformer}, Crossformer \cite{crossformer}),  non-stationary channel-clustering transformer (DUET \cite{duet}) and  CD SAM transformer (SAMformer \cite{SAMFORMER}). In addition, we also select  MLP-based models (FITS \cite{FITS}, Dlinear\cite{Dlinear}). All models are evaluated for horizons $h \in \{96,192,336,720\}$ in each dataset.  We collect the baseline results  from DUET \cite{duet} where the look-back size ($l$) is optimized for values 96, 336 and 512. To the best of our knowledge, DUET is the latest state-of-the-art work on these datasets. For fair comparison, we also run SAMformer \cite{SAMFORMER} with L1 loss and a random seed of 2022.

\subsubsection{Metrics}
In alignment with previous work, we employ mean squared error (MSE) and mean absolute error (MAE) as evaluation metrics for all models. MSE imposes a quadratic penalty  on prediction errors. Thus, a low MSE value indicates that, collectively, the predictions do not deviate significantly from the expected output. In contrast, a low MAE score indicates robust performance  with sudden changes in data, especially for datasets characterized by numerous outliers\cite{MAE_metric}. 
\subsubsection{Model Hyperparameters}
We fixed the batch size at $32$, look-back at $512$, and random seed at $2022$, to support the reproducibility of the results. The embedding dimension $d_m$ is fixed at $16$ for CCDM and we choose between $8$ and $32$ for TDM. The transformer in both modules has a single attention head. Meanwhile, the number of dependency matrix splits  $n_t = 4$ for TDM and  $n_t \in \{1, 3, 7, 20\}$ for CCDM.  We use patch sizes that are multiples of 4 or 6, such that $k \in \{6, 12, 36, 40, 44, 64, 128, 168\}$ and the random patch drop number  $n_d \in \{0, 1, 2, 3, 4\}$. The parameters of the sparsification network were initialized with samples from a normal distribution $\mathcal{N}(0, 0.5)$.
\subsubsection{Implementation Details}
All SETTer experiments are implemented in Tensorflow \cite{tensorflow2015-whitepaper} and executed on a single NVIDIA A100 GPU with 40 GB memory.
We follow the unified long-term forecasting and evaluation settings in TFB \cite{TFB}. Specifically, to ensure a fair comparison during evaluation, we do not apply the \emph{drop-last} setting to the remainder of batch sampling. The training uses L1 loss and the ADAM optimizer, with a cosine annealing warmup that transitions to exponential learning rate decay as the training progresses.

\subsection{Main Results}
Table \ref{tab:multivariate-forecasting-results2} presents SETTer results along with the baselines. To determine the best (1st count) performance among the models, the results with the same value to the nearest 3 decimal places are  highlighted in bold, and the \enquote{second best} is underlined for such cases. We outline our findings as follows.
\subsubsection{Improvements over state-of-the-art}
Overall, SETTer has the best performance. It leads the \enquote{1st count} category in $84\%$  of the test settings for MSE and  $75\%$  for MAE. The second best is DUET, with $22\%$ and $34\%$ in MSE and MAE, respectively. Furthermore, across all datasets, SETTer records an average improvement of $4.8 \%$ in MSE on $100 \%$ of the datasets and an improvement of $0.8 \%$ in MAE on $88 \%$ of the datasets, compared to DUET. 

\subsubsection{Adaptability to varying channel dimensionalities and dependence complexities} SETTer consistently outperforms CI models (e.g. PDF, pathformer, PatchTST) on all ETT datasets; although, the low channel size and weak channel correlation should favor the CI models. However, other CD models, such as iTransformer, SAMformer, and Crossformer, struggle in the same setting. SETTer outperforms Itransformer by $9.5\%$ in MSE and $6.6\%$ in MAE, SAMformer by $6.8\%$ in MSE and  $3.6\%$ in MAE and Crossformer by $43.3\%$ in MSE and  $30.3\%$ in MAE, respectively. This result indicates that SETTer's flexible auxiliary structures provide a significant advantage over existing CD methods for datasets with weak channel correlation.

SETTer also maintains its performance on datasets with larger channel dimensionality (Electricity and traffic), where complex inter-channel dependencies are common. It outperforms Itransformer by $2.3\%$ in MSE and $6.7\%$ in MAE, SAMformer by $9.5\%$ in MSE and  $7.6\%$ in MAE, and Crossformer by $21\%$ in MSE and  $7.8\%$ in MAE. It also outperforms DUET which combines CI and CD by $0.7\%$ in MSE and $0.2\%$ in MSE, highlighting the benefits of the TDM and CCDM strategy.

\subsection{Interpretable Representation Analysis}
The decoupled fixed (\emph{FMP}) and adaptive (\emph{AM}) attention computations in \ref{equation: temp self-att} and  \ref{equation: channel attention weight} make SETTer suitable for representation analysis. We decompose SETTer into its constituent temporal and channel attention components without retraining, and inspect their relative effects on LMTF. This in turn provides some interpretation of the discriminative pattern of SETTer. Specifically, we independently mask the components during forward pass and record the output, Table \ref{tab:repAnalysis} presents the average MSE over all horizons for each dataset after masking components. 
\begin{table}[H]
\caption{SETTer result with attention decoupling.}
\label{tab:repAnalysis}
\tiny
\center
\begin{tabular}{lllllllll}
\toprule
  & \textbf{ETTh1} & \textbf{ETTh2} & \textbf{ETTm1} & \textbf{ETTm2} & \textbf{Weather} & \textbf{Elec.} & \textbf{Solar} & \textbf{Traffic} \\
\midrule
\emph{Full} & \textbf{0.392} & \textbf{0.329} & \textbf{0.336} & \textbf{0.245} & \textbf{0.217} & \textbf{0.155} & \textbf{0.187} & \textbf{0.391} \\
\midrule
$\textit{AM}_T $  & 0.399 & 0.331 & 0.342 & 0.263 & 0.222 & \textbf{0.155} & 0.188 & 0.394 \\
$\textit{FPM}_T$  & 0.398 & 0.339 & 0.348 & 0.254 & 0.238 & 0.161 & \textbf{0.187} & 0.393 \\
$R_{T_{50}}$ & \textbf{0.392} & \textbf{0.329} & \textbf{0.336} & \textbf{0.245} & \textbf{0.217} & \textbf{0.155} & \textbf{0.187} & \textbf{0.391} \\
$R_{T_{95}}$ & 0.393 & 0.330 & 0.337 & \textbf{0.245} & 0.218 & \textbf{0.155} & \textbf{0.187} & \textbf{0.391} \\
\midrule
$\textit{AM}_C$  & 0.393 & \textbf{0.329} & 0.338 & 0.247 & 0.218 & 0.157 & 0.198 & 0.392 \\
$\textit{FPM}_C$ & 0.395 & \textbf{0.329} & 0.341 & 0.251 & 0.223 & 0.159 & 0.206 & 0.392 \\
$R_{C_{50}}$ & \textbf{0.392} & \textbf{0.329} & 0.337 & \textbf{0.245} & 0.218 & \textbf{0.155} & \textbf{0.187} & \textbf{0.391} \\
$R_{C_{95}}$ & \textbf{0.392} & \textbf{0.329} & \textbf{0.336} & \textbf{0.245} & 0.218 & 0.156 & \textbf{0.187} & \textbf{0.391} \\

\bottomrule
\end{tabular}
\end{table}
The complete (\emph{Full}) SETTer attention is a combination of the following components:
\begin{equation} \textit{Full} = \textit{FPM}_T + \textit{AM}_T + \textit{FPM}_C + \textit{AM}_C\end{equation} 
where $\textit{FPM}_T$ and $\textit{AM}_T$ denote short- and long-term attention, respectively.  $\textit{FPM}_C$ and  $\textit{AM}_C$ denote channel-wise self-attention and inter-channel attention.  In addition, $R_{._{50}}$ and $R_{._{95}}$ denote $50\%$ and $95\%$ random masking of the adaptive components.
We summarize the main insights as follows.
\subsubsection{SETTer displays flexible discriminative  patterns}
The results on the ETT datasets suggest that SETTer consistently weights the temporal components higher in the predictions. This is evident from the minimal drop in performance from masking the channel components, in contrast to masking the temporal components. It indicates that the low dependence of SETTer on channel relationships may provide insight into its dominance over other CD methods on the ETT datasets. Notably, the relevance of the joint temporal and channel components becomes significant only in Weather and other larger dimensional datasets, where there are considerable drops in performance from masking either of the components. This is attributable to the fact that the variables in datasets such as Weather and Electricity follow vector autoregressive process with strong coupling; thus, making their predictions to not only depend on individual past values but also strongly on the evolution of other variables.
\subsubsection{SETTer exploits long-span repetitive patterns}
SETTer uniquely demonstrates similar or identical results in temporal components decoupling on Solar and Traffic datasets. The results from temporal decoupling are also similar to \emph{full}. This unique property suggests the presence of stable/long-term repetitive patterns in the datasets, which SETTer captures in both the short- and long-term attention components. We further confirm these properties in the datasets using seasonality strength analysis \cite{TFB} on all datasets, which quantifies how recurring seasonal patterns affect variance in the data. Solar and Traffic had the highest scores of $0.746$ and $0.701$, respectively. Similar characteristics are evident from channel decoupling on Traffic dataset, whereby the results indicates that SETTer exploits the higher-order channel relationships in the dataset, such that the performance minimally drops from masking either of the the channel attention components.
\subsubsection{Potential for more sparsity in SETTer}
The absence of degradation for $R_{T_{50}}$ in all datasets indicates that SETTer can further balance complexity and expressivity through more strategic sparsification in its adaptive temporal components. A similar trend is also present from random inter-channel relation masking ($R_{C_{50}}$ and  $R_{C_{95}}$).  

\subsection{ Ablation Studies}

Here, we study the effect of full-attention transformers as substitutes for the sparse architectures in SETTer. Similarly, we investigate the distinct impact of TDM, CCDM, and channel relevance for datasets with different characteristics. Table \ref{tab:ablationanalyis} presents the results and our observations are as follows.
\begin{table}[H]
\caption{Ablation studies. Results are averaged from all horizons.} 
\label{tab:ablationanalyis}
\tiny
\center
\begin{tabular}{l | cc cc cc  cc} 
\toprule
\textbf{Datasets} & \multicolumn{2}{c}{\textbf{ETTh1}} & \multicolumn{2}{c}{\textbf{ETTh2}} & \multicolumn{2}{c}{\textbf{ETTm2}} & \multicolumn{2}{c}{\textbf{Weather}}\\
 Metrics      & MSE & MAE & MSE & MAE & MSE & MAE  & MSE & MAE \\
\midrule
SETTer & \textbf{0.392} & \textbf{0.417} & \textbf{0.329} & \textbf{0.379}  & \textbf{0.245} & \textbf{0.304} & \textbf{0.217} & \textbf{0.249} \\
\midrule
Without TDM & 0.396 & \textbf{0.417} & 0.340 & 0.388 & 0.247 & 0.306 & 0.221 & 0.250 \\
Without CCDM & 0.396 & 0.418 & 0.333 & 0.380  &  0.249 & 0.305 & 0.221 & 0.252  \\
With full attention & 0.398 & 0.418 & 0.338 & 0.383 & 0.251  &0.309 &  0.221&  0.251 \\ 
Without channel relevance & 0.397 & 0.420 & \textbf{0.329} & \textbf{0.379}  &0.247  &0.306 &  0.219&  0.250 \\ 
\bottomrule
\end{tabular}
\end{table}
\subsubsection{Impact of channel relevance}
Removing  the channel relevance block leads to the most performance decrease in ETTh1, which has a weak channel correlation and a high seasonal strength. This indicates that reliance on mere channel dependence in such settings can be noisy and can substantially deteriorate performance.
\subsubsection{Strong impact of sparse  attention}
Experiments with full-attention transformers consistently degrade performance. Similarly, the full attention configuration results in the highest average increase in the metrics, which demonstrates the positive impact of sparse encoders in long-term time series modeling and forecasting. In contrast, the other configurations with sparse attention have varying impacts on the results, highlighting how wide-range variations in the characteristics of the datasets can affect modeling choices.
\subsubsection{SETTer balances temporal and channel dependencies}
SETTer outperforms all the architecture variants and avoids overfitting to dominant temporal or channel dependencies in the datasets. Combining the TCM and CCDM modules makes SETTTer robust and expressive to consistently capture the intrinsic pattern in each dataset.

\subsection{Computational Complexity Analysis}
The complexity analysis of SETTer discussed here focus solely on the transformer modules (TDM and CCDM), which have input lengths $T$ and $C$, respectively. The most expensive computation of the two modules dominates the time complexity because of their independence. Hence, the input size $N = \mathrm{max}(T, C)$. On the one hand, the fixed masking mechanism restricts the attention computation of each token to $k$ neighbors, where $k \ll N$. The adaptive masking mechanism dynamically selects  $n \ll N$ more neighbors, resulting in $\mathcal{O}(k + n)$ per token, where $(k + n) < N$. Overall, SETTer incurs a computational cost of $\mathcal{O}(N\text{log}N)$.

\section{Conclusion}
In this work, we presented SETTer, an effective sparse-encoder model for long-term multivariate time series forecasting. SETTer introduced a novel encoder structure via a decoupled attention mechanism and hybrid masking. These mechanisms provide valuable insights on the key temporal and channel dependencies in multivariate time series data, which are also consistent with information on the physical properties of the data sources. SETTer showed impressive performance in the presence of weak channel correlation, large dimensionality, distribution shifts, and strong seasonality. It produced the best overall performance compared to existing state-of-the-art mehtods. In the future, we will be excited to explore SETTer in the development of time series foundation model.
We observe that SETTer exhibits slow convergence when the channel dimension is large; therefore, improving it via enhanced optimization techniques is another promising future direction.
\bibliographystyle{IEEEtran}
\bibliography{references}

\end{document}